\documentclass{article}

\usepackage[final]{nipless}

\usepackage[utf8]{inputenc} 
\usepackage[T1]{fontenc}    
\usepackage{hyperref}       
\usepackage{url}            
\usepackage{booktabs}       
\usepackage{amsfonts}       
\usepackage{nicefrac}       
\usepackage{microtype}      
\usepackage{graphicx}
\usepackage{comment}
\usepackage{amsmath}
\usepackage[ruled]{algorithm2e}
\SetKwProg{Fn}{Function}{}{}

\title{Deep Episodic Value Iteration 
\\
for Model-based Meta-Reinforcement Learning}

\author{
  Steven S.~Hansen \\
  Department of Psychology\\
  Stanford University\\
  Stanford, CA 94305 \\
  \texttt{sshansen@stanford.edu} \\
}

\begin{document}

\maketitle

\begin{abstract}

We present a new deep meta reinforcement learner, which we call Deep Episodic Value Iteration (DEVI). DEVI uses a deep neural network to learn a similarity metric for a non-parametric model-based reinforcement learning algorithm. Our model is trained end-to-end via back-propagation. Despite being trained using the model-free Q-learning objective, we show that DEVI's model-based internal structure provides `one-shot' transfer to changes in reward and transition structure, even for tasks with very high-dimensional state spaces.

\end{abstract}

\section{Introduction}

The deep reinforcement learning paradigm popularized by [1] has been uniquely capable of obtaining good asymptotic performance in complex high-dimensional domains. By using highly non-linear function-approximators (e.g. deep neural network), traditional model-free reinforcement learning algorithms can solve highly non-linear control problems. As this work has matured, interest has shifted from asymptotic performance to sample complexity and task transfer. It is well known that model-free methods provide poor transfer, as the knowledge of the transition and reward functions are "baked in" to the representation of the value function. 

Meta-learning has begun to appear as one possible solution the issues of sample complexity that deep learning models suffer from. By treating `solve this task from few examples' as an instance of a meta-task, slow gradient based learning is only applied across tasks, with within-task learning delegated to the forward pass of the network. While traditional networks architectures could do this in principle, much of the current success comes from networks designed to parametrize a specific nonparametric algorithm [2]. However, most previous attempts have focused on supervised learning, especially few-shot classification. Recently, there have been a few attempts at meta reinforcement learning, but none utilize the non-parametric internal structures that have proven vital in the classification context [15,16].

\section{Background}

While reinforcement learning as a field has been around for decades, with a rich variety of algorithmic approaches, most of the work utilizing non-linear function approximation (e.g. deep neural networks) had focused on the model-free algorithm Q-learning. In this section we review this algorithm alongside value iteration and kernel-based reinforcement learning, the model-based reinforcement learning algorithms that forms the foundation of our approach.

\subsection{Q-learning}

Consider the discrete action Markov decision process (MDP) $M:\{S,A,R,T,\gamma\}$, where $S$ is the set of all possible states (possibly continuous), $A$ is the finite set of discrete actions, $R$ is the reward function mapping state-action-next-state tuples to a scalar immediate reward, and $T$ is the state transition function, mapping state-action-next-state tuples to the probability of that transitions occurrence: $T(s,a,s') = p(s'|s,a)$, and $\gamma$ is discount factor weighing immediate versus future rewards. We wish to find the policy, $\pi(s,a) = p(a|s)$, that maximizes the expected cumulative discounted reward:

\begin{equation}
V(s)=\sum_{t=0}^{\infty} \sum_{a \in A} \gamma^t \pi(s_t,a)T(s_t,a,s_{t+1}) R(s_t,a) 
\end{equation}

The Bellman equation gives the relationship between the value of a state and the value of its successor:

\begin{equation}
V(s) = max_a \sum_{s'} T(s,a,s')( R(s,a) + \gamma V(s'))
\end{equation}

Notice that this can be rewritten for state-action values and expressed as an expectation:

\begin{equation}
Q(s,a) =  \mathbf{E}[R(s,a),+\gamma max_{a'} Q(s',a')]
\end{equation}

Because we have eliminated the explicit usage of $T$, this second equation can be evaluated without a dynamics model, simply by taking actions and observing their consequences (i.e sampling from the underlying state and reward transition functions). The Bellman equation holds only for the optimal value function. We can thus view the difference between LHS and RHS as the error in the relationship between the value estimates at time $t$ and $t+1$. Since we can sample directly from the true reward function (i.e. by taking actions in the real environment and observing the outcomes), it makes sense to treat the RHS as the target value for the LHS. After making this value symmetric, we can use it as a loss function for training a parametric function approximator such as a deep neural network:

\begin{equation}
    loss = \mathbf{E}[(Q(s,a) -(r(s,a) + \gamma max_{a'} Q(s',a')))^2]
\end{equation}

Optimizing this temporal difference loss using stochastic gradient descent is non-trivial, as iid sampling from a MDP is non-trivial, and there is a dependency between the current estimated and target values. Both problems are addressed in [1], using random samples from a `replay buffer' of past transition samples $\{s,a,r,s'\}$  to address iid concerns, and a slow changing `target network' to break the dependence between estimated and target values.

\subsection{Matching Networks}

\begin{figure}[h]
  \centering
  \fbox{\includegraphics[scale=.4]{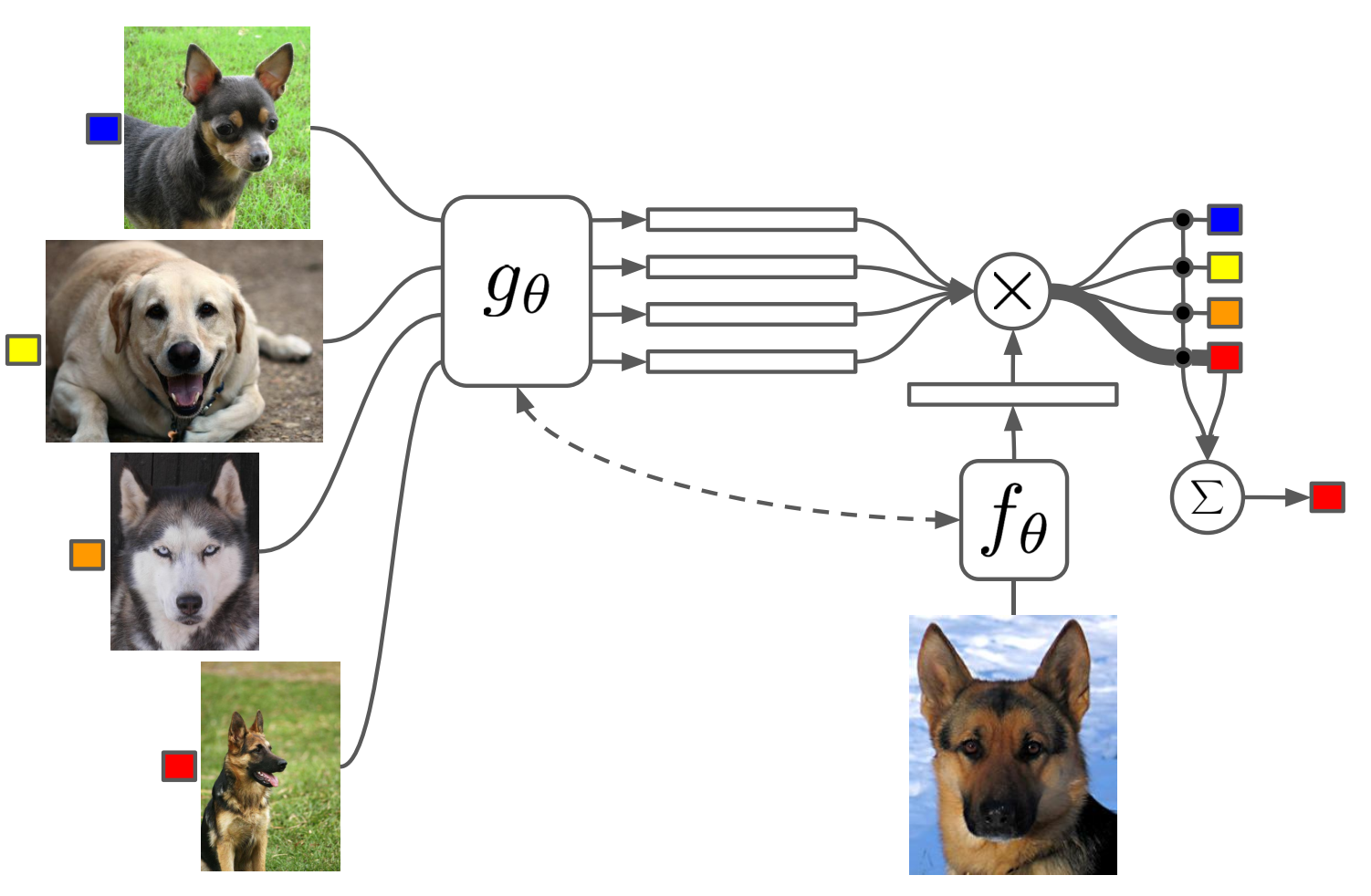}}
  \caption{The matching network architecture (reproduced with permission from [2]). A neural-network ($f_\theta$) is used to embed an query image into a latent space, alongside several other images with known labels ($g_\theta$). A similarity vector is then produced by passing the query latent vector through a cosine kernel with each known latent vector. This vector is then normalized and multiplied by the vector of known labels to produce an estimate of the query image's label. All of these operations are differentiable, so the supervised loss can be back-propagated to the parameters of the embedding network.}
  \label{fig:matching}
\end{figure}

Despite their focus on classification, the matching networks of Vinyals et al [2] are the closest prior work to our approach. As shown in figure \ref{fig:matching}, matching networks use kernel-based averaging to estimate label probabilities, and learn the latent-space over which the kernel is applied end-to-end.

Relating this approach to the reinforcement learning problem, one could imagine using a kernel to relate a novel state to states with known values (i.e. kernel weighted regression). Blundell et al do this with a nearest neighbor kernel and empirical returns as the value estimates of previously encountered states [4]. While the embedding space used by the kernel was based on either random projections or learnt in an unsupervised fashion, the kernel could theoretically be learned in an end-to-end manner based on a loss between predicted and actual returns. Episodic Neural Control [14] recently made something akin to this extension, and will be explored in greater detail in the discussion section. However, both works were constrained to deterministic MDPs with frequently revisited states, limiting its overall applicability.

\subsection{Kernel-based Reinforcement Learning}

Kernel-based reinforcement learning (KBRL) is a non-parametric reinforcement learning algorithm with several advantageous properties [5]. It is a model-based reinforcement learning algorithm, but relies on stored empirical transitions rather than learning a parametric approximation. This reliance on empirical state and reward transitions results in low sample complexity and excellent convergence results.  \footnote{Convergence to a global optima assuming that underlying MDP dynamics are Lipschitz continuous, and the kernel is appropriately shrunk as a function of data. While this result doesn't hold when a non-linear function approximator is used to adapt the kernel, it seems to converge empirically with proper hyper-parameter tuning.}.

The key insight of this algorithm is that a set of stored transitions $D_A = \{S,R,S'\}$ can be paired with a similarity kernel  $\kappa(x_0,x_1)$ to represent each resultant state ($s' \in S'$) as a distribution over the origin states ($S$). This new MDP can approximate the real one arbitrarily closely (as the stored transitions go to infinity and the kernel bandwidth goes to 0), but has the advantage of being closed under states whose transitions have been sampled from the environment. 

To see why this is the case, note that, empirically, we know what the state transitions are for the stored origin states:

\begin{equation}
 \forall s \in S :
    \begin{cases} 
      T_a(s,s') = 1/\sum\{s,\cdot,\cdot\}\in D_a & \{s,\cdot ,s'\} \in D_a \\
      T_a(s,s') = 0 & otherwise
   \end{cases}
\end{equation}

Likewise we know the reward transitions for these states:

\begin{equation}
\forall s \in S :
      r_{s,a} = \mathbf{E} [R_a(s)]
\end{equation}

Since we are dealing with a continuous space, the state resulting from one of these transitions is unlikely to have its own stored transition (i.e. $\exists s' \in S' : s' \notin S$). Using the similarity kernel, we can replace this unknown resultant state with a distribution over the origin states. This makes the state transitions from $S \rightarrow S$ instead of $S \rightarrow S'$, meaning that all transitions only involve states for which we have experienced empirically.

Since there are a finite number of states (equal to the number of stored transitions), and these states have known transitions, we can perform value iteration to obtain value estimates for all resultant states $S'$ (the values of the initial states $S$ aren't needed, as the bellman equation only evaluates states \emph{after} a transition). We can obtain an approximate version of the bellman equation by using the kernel to compare all successor states to all initial states in the transition set:

\begin{equation}
\forall x \in S' :V(x)\leftarrow max_a \sum_{s\in S_a} \kappa(x,s)[R_a(s)+\gamma V(s_a')]
\end{equation}

\section{Model} 

The KBRL algorithm is great in theory, but suffers from the so-called `curse of dimensionality' -- as the dimensionality of the state space grows, the number of exemplars needed to cover the space grows exponentially. By taking the KBRL algorithm and making the similarity kernel act on the output of a deep neural network, we can force the data to live in a lower dimensional space. While random projections could accomplish the same thing, an embedding function optimized on overall performance should be able to do a much better job, by focusing on performance relevant aspects of the observed states.

\begin{algorithm}
\DontPrintSemicolon
\caption{Deep Episodic Value Iteration}
$\kappa(\hat x,x_i) = e^{cosine(\hat x,x_i)}/\sum^k_{j=1} e^{cosine(\hat x,x_j)} $\;
\Fn{Q ($s,a,E,limit;\theta$)}{
$(S,R,S') := E$ \;
$Z := Encoder(S;\theta)$ \;
$Z' := Encoder(S';\theta)$ \; 
$z := Encoder(s;\theta)$ \;
\For{$a := 1, A$}{
    $\Theta_a := \kappa(Z',Z_a)$
}
$V := 0$\;
\For{$i := 1,limit$}{
    $V := max_a \Theta_a [R_a+\gamma V_a]$
}
\Return{$\kappa(z,Z_a)[R+\gamma V]$}
}
Initialize replay memory $D$ to capacity $N$ \;
Initialize action-value function $Q$ with random weights \;
\For{$episode := 1, M$}{
\For{$t := 1, T$}{
\For{$a := 1, A$}{
    Sample random set of transitions $E_a := (s_{a_k},r_{a_k},s_{a_{k+1}})$ from $D(\cdot,a,\cdot,\cdot)$\;
    }
With probability $\epsilon$ select a random action $a_t$\;
otherwise select $a_t := max_aQ(s_t,a,E,T;\theta)$ \;
Execute action $a_t$ and observe reward $r_t$ and observation $s_{t+1}$ \;
Store transition in $(s_t,a_t,r_t,s_{t+1})$ in $D$ \;
Sample random minibatch of transitions $(s_j,a_j,r_j,s_{j+1})$ from $D$ \;
$loss := 0$ \;
\For{$i := 1, T$}{
\uIf{terminal or $i = 1$}{$y_j := r_j$}
\Else{$y_j := r_j+\gamma max_{a'} Q(s_{j+1},a',E,i-1;\theta)$}
$loss_j := loss_j + (y_j-Q(s_j,a_j,E,i;\theta))^2$ 
}
Perform a gradient descent step on $loss$ \;
}
}
\label{algo:main}
\end{algorithm}

The resulting model, Deep Episodic Value Iteration (DEVI) is KBRL applied to a state representation learned through Q-learning. This representation is the output of a deep neural network, with weights tied such that query states are encoded using the same function as states in the episodic store. To ensure that, during learning, this parametric encoder doesn’t rely upon any particular states being the episodic store, a random subsample of the episodic store is used for any given training step. In order to keep GPU memory usage relatively constant during training, the size of the subsampled set of experiences is held constant, though there is no fundamental objection to using a growing set as per KBRL.

The size of the computation graph grows linearly with the planning horizon, so DEVI uses a fixed number of value iteration steps rather than iterating to convergence (though again, this is a pragmatic rather than fundamental deviation from KBRL). For the simple tasks considered here, it was trivial to determine the maximum number of value iteration sweeps required to compute the value function, but adaptive computation techniques, such as those designed of recurrent neural networks [7], should be able to automate this process on more demanding environments.

DEVI is trained using Q-learning (modified to account for the fixed planning horizon), and all of the improvements made since the influential DQN paper (e.g. Double DQN [8]) [1] could also be incorporated. However, one unique problem that DEVI faces is the long gradient path corresponding to the value iteration steps. The structure of the computational graph means that the paths corresponding to shorter term predictions (e.g. the first value iteration step simply predicts the maximum immediate reward) are the last to receive gradient information. It is possible to eliminate this problem by using the values after each step of value iteration (rather than just the after the final step) to calculate the predicted value of the current state-action pair. Thus, if K steps of value iteration are performed, the model has K different predictions for the value of Q(s,a), corresponding to the K different planning horizons. Each of these is amenable to Q-learning (using target values derived from the immediate reward and the values from the previous planning step), and these K losses are simple averaged to calculate the net loss.

\section{Initial Experiments}

The primary objective of this work is to demonstrate that `one-shot learning' is possible on reinforcement learning problems using deep neural networks. \footnote{Our definition of one-shot learning is the ability to solve a novel task without parameter updates (i.e. changes to the weights of the deep network). The episodic store must still be populated with transitions from the new task, but this requires orders of magnitude fewer samples than any parametric adaptation.} As there has been no prior work demonstrating this phenomenon, we've designed a toy task domain as a proof of concept -- it is as minimalistic as possible while still allowing for a variety of tasks requiring non-linear state representations.

\subsection{Omniglot Graph World}

\begin{figure}[h]
  \centering
  \fbox{\includegraphics[scale=.33]{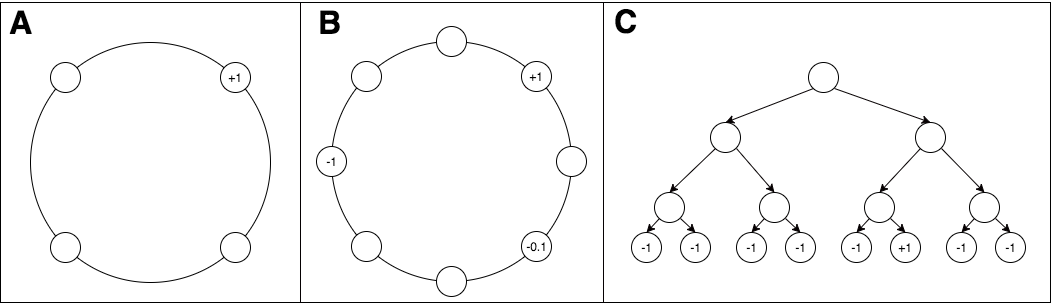}}
  \caption{Three sub-domains within the Omniglot graph world. The nodes represent the states and the edges representing the connectivity resulting from the possible actions. The agent observes images of characters from the Omniglot dataset rather than the states themselves, with this mapping being randomly drawn for each new task. Additionally, the relative reward locations and action semantics are permuted. A) The Ring task distribution consists of a few states (10 in the experiments) and a single positive  terminal reward. B) The Hard Ring task distribution consists of more states (15 in the experiments), large terminal negative reward and a small non-terminal reward. C) The Tree task distribution has states that form a full binary tree (depth 5 in the experiments), with all leaf nodes except one giving negative terminal rewards.}
  \label{fig:omniglot}
\end{figure}

The Omniglot dataset consists of 28 by 28 pixel grey-scale images depicting characters from real and fictional languages. Often referred to as the `transpose of MNIST', there are 1600 different classes of character, but only 20 examples per class. Due to this limited sample size, it has quickly become the standard bearer for `one-shot' learning [19]. However, the format of the dataset is only conducive to `one-shot' learning for classification. Here we extend the dataset for use as a proof-of-concept for `one-shot' reinforcement learning.

The Omniglot Graph World is a task domain consisting of all MDPs with a small finite number of states, each randomly assigned to a character class from the Omniglot dataset. Rather than observe the states directly, the agent only has access to a sample from the corresponding Omniglot class.

While trivially easy to solve in isolation, the performance criterion of interest is the number of samples required to learn a new task in this domain, given prior exposure to some other tasks. This domain is interesting in that all of the tasks share the same similarity structure (i.e. observations coming from the same class are bisimilar), while differing greatly in their state and reward transition structure.

The number of tasks contained within the Omniglot Graph World is huge and vary widely in difficulty due to the possibility of degenerate transition structures (e.g. all states equally rewarding). So for the purposes of thorough evaluation, we've constrained the space of MDPs to variations on 3 prototypes (as shown in Figure \ref{fig:omniglot}), each having 2 actions and unique transition structures.

\subsection{Implementation Details}

Analogously to work on `one-shot' classification, we trained DEVI in an interleaved manner across tasks randomly sampled from our task domain. Rather than interleave within mini-batches, tasks were switched out after every step, and $1000$ `burn in' steps were taken to fill in a small replay buffer from which to sample a minibatch of $100$ tuples for Q-learning. \footnote{Initial experiments showed that longer gaps between interleaving didn't alter performance, so episode level interleaving should be used to avoid intractability on larger tasks.}

DQN was trained for $1000$ minibatches on both the initial and transfer tasks. A replay buffer was used with a capacity of $100000$, and was filled prior to the start of learning to avoid tainting our sample complexity analysis.

To focus the evaluation on between task transfer rather than within task exploration, we decided to train all models with a uniform random behavior policy. This ensures that the data provided to the different algorithms is equivalent, and performance differences aren't due to more structured exploration. \footnote{while intelligent exploration isn't a free byproduct of our approach, it is easily obtained by exploiting the pseudo-tabular nature of the learned model. An extension of R-MAX, whereby neighbors within a fixed distance substitute for visitation counts, has been shown by Li et al [9] to be quite effective.} We also modified DEVI's episodic store such that it always contains 5 samples from each state. Randomly sampling from the replay buffer (as described in Algorithm \ref{algo:main}) works similarly, but this method allows for useful comparison with the work on `one-shot' learning for classification.

The same basic deep neural network architecture is used for both algorithms, taken directly from the Matching Networks paper [2]. It consists of four $3 \times 3$ convolutional layers, each with 64 filters, interleaved with $2 \times 2$ max-pooling. For DQN, this is followed by a densely connected layer ($100$ units) with ReLU activations. This is used to encode the latent space in DEVI, and to represent the value function in DQN. However, since the output dimensionality of DQN is tied to the number of actions, a final linear layer is added.

The ADAM optimizer was used for both algorithms with a batch-size of $100$. Default hyper-parameters were used, though a preliminary grid search showed that these didn't significantly impact performance across a wide range of values. Likewise, extensions to DQN were considered (e.g. Double DQN), but none yielded qualitatively different results.

\begin{figure}[h]
  \centering
  \fbox{\includegraphics[scale=.75]{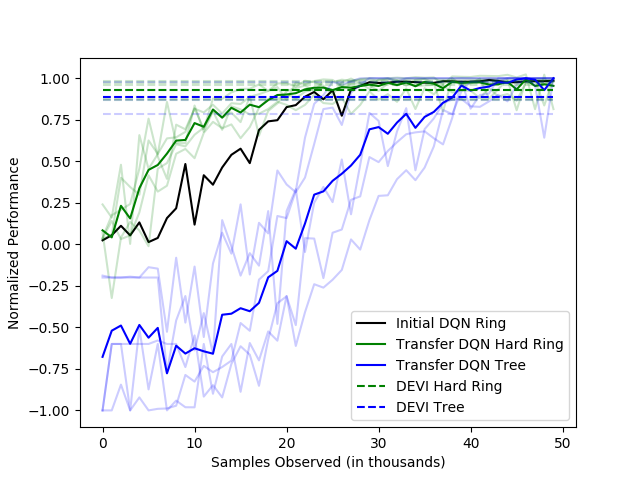}}
  \caption{results on the Omniglot Graph task domain. 5 instances of DQN and DEVI were initially trained using random seeds. The average performance over time for this initial phase is shown in black for DQN. DEVI was trained over a distribution of tasks, so its performance curve is not shown. Each of the 5 models were then used as the initial weights for transfer learning. 5 transfer attempts were performed for each model and averaged together, with the average across models in bold. In the case of DEVI, these transferred weights were frozen to emphasize the non-parametric nature of their generalization.}
  \label{fig:results}
\end{figure}

\subsection{Results}

As Figure \ref{fig:results} shows, DEVI was able to instantly generalize to new tasks, even though none the characters representing the underlying states had ever been seen before (they were drawn from a held out set of characters). In addition, both test task distributions had qualitatively different state and reward transition structures from that of the training task distribution. As expected, DQN exhibited very little transfer, needing samples on the same order of magnitude as networks initialized with random weights. 

The differences between these two models give evidence to the claim that while both are trained using Q-learning, a model-free reinforcement learning algorithm, DEVI's non-parametric internal structure allows for learning representations that work across a wide array of tasks. More precisely, this structure paired with interleaved learning forces DEVI to learn an appropriate latent space for kernel-based planning.

\section{Discussion}

These initial experiments demonstrate the possibility of meta model-based reinforcement learning, but it remains unclear how well this approach will scale to realistic domains. There have recently been several papers proposing model-free meta reinforcment learning methods [14,15,16], and it is an interesting open question how DEVI's model-based generalization compares to fast model-free adaptation.

\subsection{Related Work}

The most similar among this recent work on model-free meta reinforcement learning is Neural Episodic Control [14]. Interestingly, it isn't presented as a meta learning algorithm, but rather as just sample-efficient learner. Despite this, NEC can be seen as the offspring of Matching Networks, as it uses the same kernel-based averaging, but to weight values instead of labels. It shares with DEVI the idea of training the encoding network using Q-learning, but differs in only storing the latent space in the episodic store, meaning that the encoder is only updated for the queried state, not the states it is being compared to. Instead, the latent vectors are themselves treated like parameters once stored and modified via back-propagation. Additionally, similarity was approximated by only comparing a query against its $k$ approximate nearest neighbors. These two tricks allow NEC to scale to very complex tasks such as the Atari 2600 environment.

Despite the similar titles, `Value-Iteration Networks' [11] and this work are related, but quite distinct. In that work, the authors assumed a underlying 2D state-space with local transitions. This allowed them to perform efficient value iteration by noticing that for this specific topology, the value iteration algorithm can be performed by 2D convolution. Additionally, this approach made the reward and value function internal (end-to-end training was performed from the resulting \emph{policy}), foregoing any attempt to learn a veridical reward function. The work presented here shares neither of these qualities, though it is an interesting open-question as to which approach is more likely to scale to complex domains.

\subsection{Future Directions}

One pragmatic concern is that this architecture scales poorly as the size of the episodic store grows. In the cognitive science literature, it is well known that fixed episodic stores can be augmented by traditional neural networks to incorporate knowledge from the distant past [13], though this has yet to be empirically validated on a large scale model. A more immediate (and complementary) solution would be to sparsify the similarity matrices by zeroing out all but the nearest $k$ neighbors for each row. While this could theoretically make gradient propagation more problematic, recent related work on Neural Episodic Control and sparsified Neural Turing Machines suggests that this approach can be quite effective empirically  [14,10].

Due to DEVI being both similarity-based and model-based, it is worth emphasizing that even without interleaved learning of multiple tasks, significant generalization should be possible. So long as the optimal Q-values of the initial task don't alias state-action pairs that are important to a transfer task, then `one shot' transfer should be possible to the extent the two tasks share similarity structure. For example, changing reward structures in a fixed environment would guarantee that only value aliasing would prevent transfer. Work on the so-called Successor representation fleshes out a similar argument, though its architecture imposes the harsher limitation of \emph{immediate} reward aliasing (which is significant for the sparely rewarded environments common to the deep reinforcement learning literature), and it has additional constraints owing to its on-policy nature [17,18].

Recently, researchers have noted that humans' ability to react to changes in task structure far surpasses that of deep reinforcement learning approaches such as DQN [12]. Specifically, these authors note that even extreme alterations to reward structure (e.g. lose an Atari game instead of winning), are readily solvable by humans but not deep learning systems. As previously mentioned, our model is unique in being able to rapidly adjust to changes in reward structure. While there are several technical hurdles to overcome, we believe that DEVI will be able to tackle this challenge within the immediate future.

\section*{References}

\small

[1] Mnih, Volodymyr, et al. "Human-level control through deep reinforcement learning." {\it Nature} 518.7540 (2015): 529-533.

[2] Vinyals, O., Blundell, C., Lillicrap, T., \& Wierstra, D. (2016). Matching networks for one shot learning. {\it In Advances in Neural Information Processing Systems} (pp. 3630-3638).

[3] Santoro, A., Bartunov, S., Botvinick, M., Wierstra, D., \& Lillicrap, T. (2016). Meta-learning with memory-augmented neural networks. {\it In Proceedings of The 33rd International Conference on Machine Learning} (pp. 1842-1850).

[4] Blundell, C., Uria, B., Pritzel, A., Li, Y., Ruderman, A., Leibo, J. Z., ... \& Hassabis, D. (2016). Model-free episodic control. {\it arXiv preprint arXiv}:1606.04460.

[5] Ormoneit, D., \& Sen, Ś. (2002). Kernel-based reinforcement learning. {\it Machine learning}, 49(2-3), 161-178.
Chicago	

[6] Dayan, P. (1993). Improving generalization for temporal difference learning: The successor representation. {\it Neural Computation}, 5(4), 613-624.
Chicago	

[7] Graves, A. (2016). Adaptive computation time for recurrent neural networks. {\it arXiv preprint arXiv}:1603.08983.

[8] Van Hasselt, H., Guez, A., \& Silver, D. (2016, March). Deep Reinforcement Learning with Double Q-Learning. In {\it AAAI} (pp. 2094-2100).

[9] Li, L., Littman, M. L., \& Mansley, C. R. (2009, May). Online exploration in least-squares policy iteration. {\it In Proceedings of The 8th International Conference on Autonomous Agents and Multiagent Systems-Volume 2} (pp. 733-739). International Foundation for Autonomous Agents and Multiagent Systems.

[10] Rae, J., Hunt, J. J., Danihelka, I., Harley, T., Senior, A. W., Wayne, G., ... \& Lillicrap, T. (2016). Scaling Memory-Augmented Neural Networks with Sparse Reads and Writes. {\it In Advances In Neural Information Processing Systems} (pp. 3621-3629).

[11] Tamar, A., Wu, Y., Thomas, G., Levine, S., \& Abbeel, P. (2016). Value iteration networks. {\it In Advances in Neural Information Processing Systems} (pp. 2154-2162).

[12] Lake, B. M., Ullman, T. D., Tenenbaum, J. B., \& Gershman, S. J. (2016). Building machines that learn and think like people. {\it arXiv preprint arXiv}:1604.00289.

[13] Kumaran, D., Hassabis, D., \& McClelland, J. L. (2016). What learning systems do intelligent agents need? complementary learning systems theory updated. {\it Trends in Cognitive Sciences}, 20(7), 512-534.

[14] Pritzel, A., Uria, B., Srinivasan, S., Puigdomènech, A., Vinyals, O., Hassabis, D., ... \& Blundell, C. (2017). Neural Episodic Control. {\it arXiv preprint arXiv}:1703.01988.

[15] Wang, J. X., Kurth-Nelson, Z., Tirumala, D., Soyer, H., Leibo, J. Z., Munos, R., ... \& Botvinick, M. (2016). Learning to reinforcement learn. {\it arXiv preprint arXiv}:1611.05763.

[16] Finn, C., Abbeel, P., \& Levine, S. (2017). Model-Agnostic Meta-Learning for Fast Adaptation of Deep Networks. {\it arXiv preprint arXiv}:1703.03400.

[17] Dayan, P. (1993). Improving generalization for temporal difference learning: The successor representation.{\it Neural Computation}, 5(4), 613-624.

[18] Kulkarni, T. D., Saeedi, A., Gautam, S., \& Gershman, S. J. (2016). Deep successor reinforcement learning. {\it arXiv preprint arXiv}:1606.02396.

[19] Lake, B. M., Salakhutdinov, R., \& Tenenbaum, J. B. (2015). Human-level concept learning through probabilistic program induction. {\it Science}, 350(6266), 1332-1338.

\end{document}